# Large Multimodal Model based Standardisation of Pathology Reports with Confidence and their Prognostic Significance

Ethar Alzaid[1]*, Gabriele Pergola[1], Harriet Evans[2,3], David Snead[1,2,3],

Fayyaz Minhas[1]

[1] Department of Computer Science, University of Warwick, Coventry, United Kingdom.

[2] Histopathology Department, University Hospitals Coventry and Warwickshire NHS Trust, Coventry, United Kingdom.

[3] Warwick Medical School, University of Warwick, Coventry, United Kingdom.

* Corresponding author E-mail : ethar.alzaid@warwick.ac.uk;

Contributing authors:

gabriele.pergola.1@warwick.ac.uk; harriet.evans4@nhs.net; david.snead@pathlake.org;

fayyaz.minhas@warwick.ac.uk;

## ABSTRACT

Pathology reports are rich in clinical and pathological details but are often presented in free-text format. The unstructured nature of these reports presents a significant challenge limiting the accessibility of their content. In this work, we present a practical approach based on the use of large multimodal models (LMMs) for automatically extracting information from scanned images of pathology reports with the goal of generating a standardised report specifying the value of different fields along with estimated confidence about the accuracy of the extracted fields. The proposed approach overcomes limitations of existing methods which do not assign confidence scores to extracted fields limiting their practical use. The proposed framework uses two stages of prompting a Large Multimodal Model (LMM) for information extraction and validation. The framework generalises to textual reports from multiple medical centres as well as scanned images of legacy pathology reports. We show that the estimated confidence is an effective indicator of the accuracy of the extracted information that can be used to select only accurately extracted fields. We also show the prognostic significance of structured and unstructured data from pathology reports and show that the automatically extracted field values significant prognostic value for patient stratification. The framework is available for evaluation via the URL: https://labieb.dcs.warwick.ac.uk/.

# 1 INTRODUCTION

The healthcare industry generates a vast amount of clinical data that plays a pivotal role in the provision of patient care and medical decision making, facilitating accurate and personalised treatment decisions. Therefore, the availability and precision of clinical information is fundamental to contemporary medical practice and significantly influences the quality of healthcare services. However, the unstructured nature of healthcare data is a significant and ongoing challenge [1]. Pathology reports for instance, vary in formats, styles, and terminology depending on the regulations of the health centre that generates them which makes structuring and standardisation beneficial in multiple ways. Standardisation can serve as a method for assessing reporting quality and ensuring adherence to reporting guidelines, encouraging higher quality data collection by pathologists when reporting specimens. Additionally, structured data can be used to develop advanced machine learning models for diagnosis and treatment planning. Moreover, reports uniquely contain some of the critical and prognostic features of cancer that can be useful for building models to process histology images.

Early efforts to gain insights from pathology reports applied Natural Language Processing (NLP) techniques to extract useful information [3,4,5]. Employing conventional NLP approaches can be challenging due to the complex training processes. Additionally, these models often struggle with generalisability where the performance can drop across datasets or domains due to variations in language, context, and terminology.

Zero-shot and in-context learning capabilities in Large Language Models (LLMs) offer promising solutions to these challenges. They enable models to understand and generate responses in previously unseen contexts without the need for extensive retraining or domain-specific data. Many researchers in computational pathology have made efforts for experimenting with LLMs. Generative Pre-trained Transformer (GPT-3.5) was prompted to extract clinical data from pathology reports single-line response [6] and in a questionnaire-based prompting [7].

GPT-4 demonstrated exceptional potential across numerous tasks [8]. The model shown a great potential in the medical field by passing a standardised test without fine-tuning or special prompt crafting [9]. Being a Large Multimodal Model (LMM), GPT-4 is capable of handling data of various modality, including text and images. GPT-4 zero-shot capabilities in structuring pathology reports were highlighted in [10]. They explored multiple prompting techniques, including having the model fill out a pre-defined template with extracted data and prompting the model to suggest a structured template for reporting.

All aforementioned methods developed different strategies to use LLMs for extracting clinical information from unstructured text reports. However, they do not explicitly assign their extractions with confidence scores and limiting the reliability of the output. Another limitation is prompting LLMs to extract all information at once without providing sufficient context, allowing the model greater freedom to generate undesirably creative responses. In addition, some approaches lack compliance to an established reporting standards, which defies the goal of standardisation. Some approaches may not be designed to effectively process or interpret image-based data and only take text reports as input. Also, to the best of our knowledge, no existing approach has analysed the prognostic significance of existing reports.

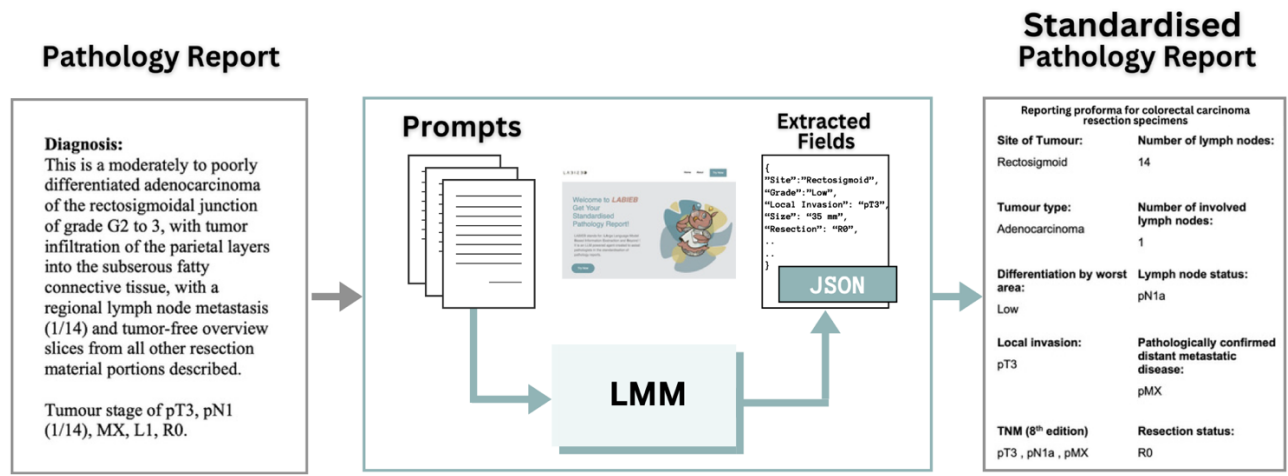

**Figure 1.** Standardisation of unstructured text-based pathology report with the proposed Large Multimodal Model (LMM) framework (LABIEB). It takes as input pathology reports and a series of structured prompts to a LMM (GPT in this case). It then extracts specific fields from the responses and reformats the extracted information into a standardised format. The website is accessible at: https://labieb.dcs.warwick.ac.uk/ .

This study aims to address these limitations by proposing a two stage information extraction structure using Large Multimodal Model (LMM). The basic concept of this work is illustrated on Figure 1. This architecture enables the estimation of a confidence score to reflect the accuracy of the extracted information. Expressing uncertainty gives the user of the data the liberty to reject samples of data with lower confidence. We used LMM over LLM to address the challenges that can pose in the first phase of extracting text from image-based pathology reports. We have used multiple prompts with adequate context to improve the overall performance of the model [11] and aid in the confidence estimation. The output of the model follows an established pathology reporting standards by the Royal College of Pathologists (RCPath) [12]. The proposed structure is tested on pathology reports for Colorectal Adenocarcinoma (COAD) patients from The Cancer Genome Atlas (TCGA) dataset. The major contributions of the paper are as follows:

1. A two-stage architecture for information extraction from scanned unstructured pathology reports using large multimodal language models (LMMs).
2. Confidence assignment to extracted fields during both stages of extraction and validation based on estimated accuracy of extracted fields.
3. Analysis of prognostic value of structured and unstructured report embeddings and fields.
4. Open availability for pathologists' use through a publicly available website.
5. Standardised reports for TCGA are made available for public use.

## 2 MATERIALS AND METHODS

The architecture for the proposed model is shown in Figure 2. The methodology of the proposed model involves two-stage prompting to two LMM agents we refer to as Extractor and Validator agents. The main function of the Extractor agent is to identify and extract the value of a specific field called the query field from the input text report. For example, we can extract grade or stage or any other information directly from the report by setting the specifying the query field to be grade or stage, respectively. The Validator's role is to assess and verify the accuracy of the Extractor's output. The final output of the model is a single response of the extracted field with an estimated confidence level that reflects the model's accuracy and reliability in information extraction. Once various fields have been extracted, we can then analyse their prognostic significance.

### 2.1 Dataset

We utilised Colorectal Adenocarcinoma (COAD) cases from The Cancer Genome Atlas (TCGA) dataset (https://www.cancer.gov/tcga) to perform experiments on the proposed model. It provides a multi-centric collection of pathology reports with varying qualities and reporting styles. Out of $N = 635$ COAD reports, 36 were filtered out due to their significant low image quality. The total number of reports remaining in the experiment is $N_T = 599$. We have created a subset of manually extracted fields of $N_V = 240$ cases which is used for validation of the accuracy of the extracted fields. The validation set is created with care to be representative of the diversity among centres in TCGA.

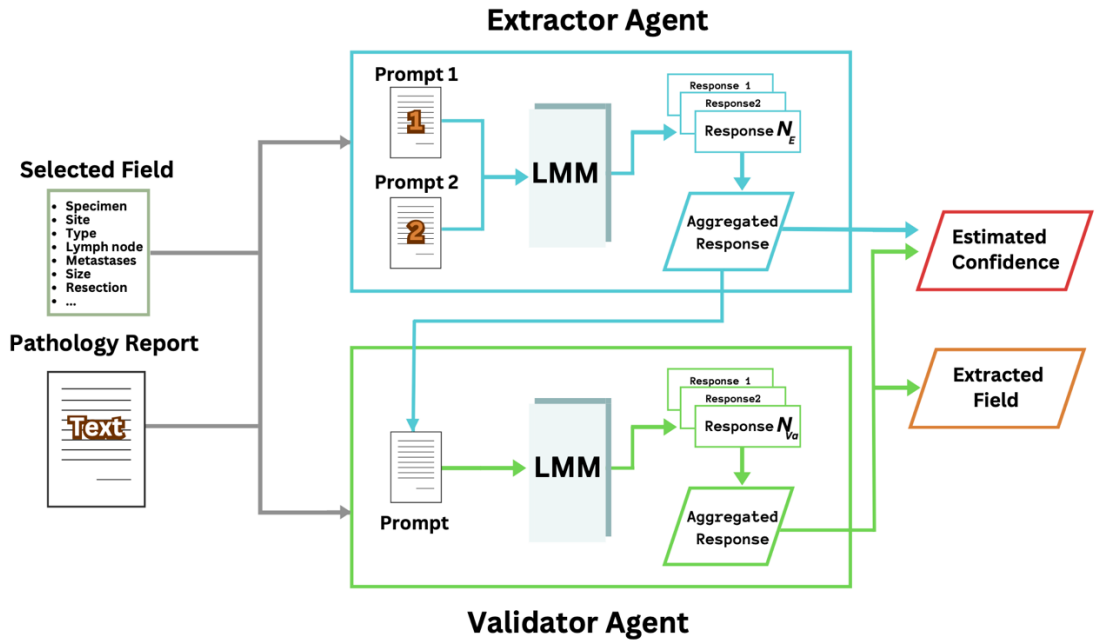

**Figure 2.** The framework for extracting information operates in two stages and takes an unstructured text report and a query field (e.g., grade or stage etc.) as input and extracts the value of query field from the report along with assignment of a confidence value to the extracted field. First the Extractor operates by incorporating the input with two prompts to be sent to the LMM to produce N_E responses which are aggregated into a single response. The same input along with the Extractor's response goes into the Validator's prompt and into the LMM. N_{Va} responses are generated and aggregated to a single response with a confidence value estimated by assessing both the Extractor's and the Validator's responses.

## 2.3 Reporting Standards

Ensuring compliance with established standards for colorectal cancer reporting is essential for standardisation and maintaining consistency across all patients' data. We have followed the Standards and Dataset for histopathological reporting of colorectal cancer by the Royal College of Pathologists (RCPath) [12]. The format of the output follows the in the RCPath standards. In addition, we have designed the prompts and aggregated the responses based on the categorisation schemes and formats of the standards. We have extracted various fields in the standards such as: Specimen type, Tumour type, Tumour site, Maximum diameter, Local invasion status, Histologic grade, Number of examined lypmh nodes, Number of metastatic nodes, Lymph node status, Distant metastatic disease status, and Resection status.

## 2.4 Text Extraction

The dataset consists of PDF reports scanned at varying image resolutions and with different textual lengths, including some with handwritten notes and others in tabular format. For these files to be utilised by LMMs, they must first be transformed into text using Optical Character Recognition (OCR). Conventional OCRs, such as the widely recognised Tesseract library [13], encountered difficulties in processing the diverse content effectively. We used the GPT-4 Vision model to address this issue as it is recognised for its strong performance with Latin-based OCR [14].

## 2.5 Extractor Agent

The Extractor is a GPT-4-Turbo powered agent aiming to extract a single field at a time from the textual report as illustrated in Figure 2. For each query field $q$, this agent sends two prompts to the LMM model to retrieve $N_E = 20$ responses in total in JSON format. The agent then returns a single response that appeared the most among the other responses. The Extractor confidence $E_{Confidence}$ is reported as the percentage of times the returned response appeared among responses for the query field. The model is instructed to respond with "Not Available" when there is no sufficient information in the report. It is possible for the model to encounter ties since it returns the query field based on how many times it appeared, the interested reader can refer to supplementary for details on this matter.

## 2.6 Validator Agent

The Validator is a GPT-4-Turbo powered agent used to validate the Extractor's response. This agent receives a curated prompt with instructions to validate the Extractor's response. For every query field, the Validator produces an output consisting of three labels: Correctness, Confidence and Corrected field value. Correctness states whether the Extractor is correct, Confidence shows how confident the model is in its response (out of 100), and Correction is the Validator's response as a correction in case the Extractor was incorrect. The LMM produces $N_{Va} = 10$ responses in total in JSON format for a single prompt and for each query field. Responses are aggregated and ties are handled with the same manner as the Extractor agent. The Validator confidence is reported as the percentage of times the returned response appeared among responses for each output label and each query field.

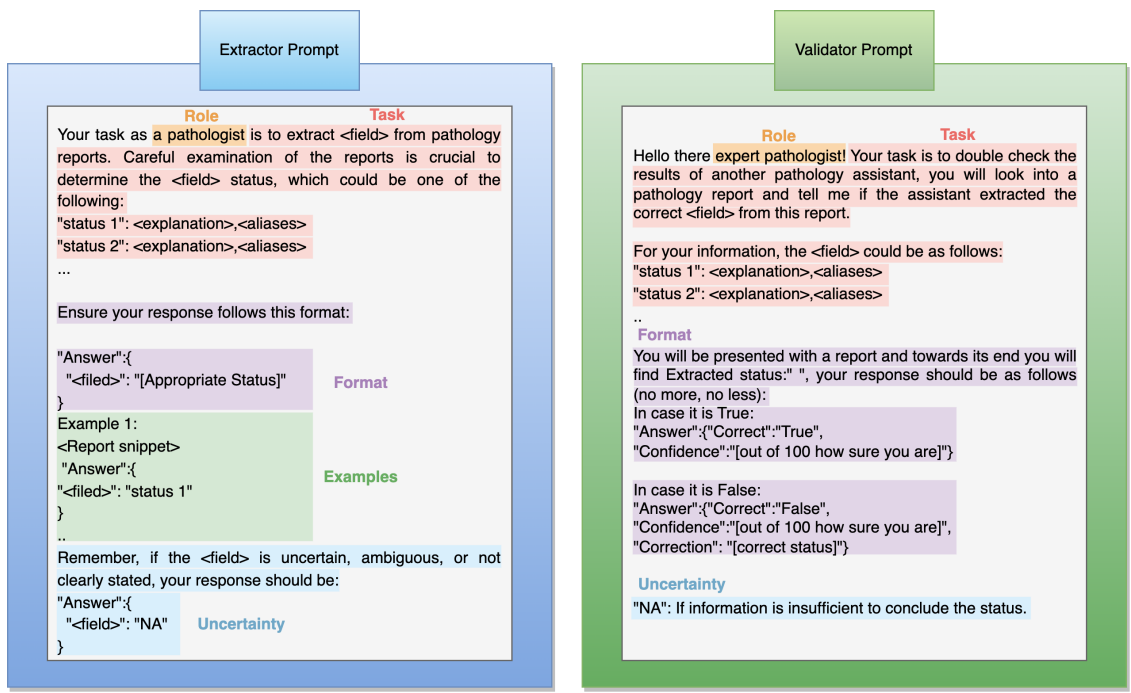

**Figure 3.** Basic structures of the Extractor and the Validator agents' prompts.

## 2.7 Prompt Design

The prompt for information extraction is composed of five main components: Role, task, format constraints, examples, and uncertainty handling. Two variations of designs are shown in Figure 3 with samples of both the information extraction and validation prompts. For each query field, we have designed two prompts for the Extractor and one for the Validator to utilise LMM's sensitivity to prompt format in confidence estimation [15].

### 2.7.1 Role and Task Specifications

The role of the agents is specified as pathologists with the task of extracting or validating a specific field from a pathology report. We explain the task in details reflecting the RCPath guidelines in the categorisation of each field. Each category is described along with its variations or aliases that the model might encounter while analysing the text for field extraction. For instance, the stage "pT4a" is explained as "Tumour cells breaching the serosa or perforating" and may also be referred to as "T4A", "pT4", or "T4".

### 2.7.2 Response Format

We have defined the model's response format and instructed the model to follow this format when it generates the response. The specified output or response format is set to JSON where the query field is stated and then followed by the generated response. Specifying the format reduces output formatting errors and this reduction can be attributed to the fact that outlining the response structure limits the model's hallucination probability and directs its efforts on the specified task [16].

### 2.7.3 Example Scenarios

We have provided the model with a set of example scenarios for each query field where each example includes a segment of a report and the response we expect the model to generate. For instance, in the local invasion prompt, we have given a segment of a report stating ("Tumor invades muscularis propria,") followed by the expected response ("Local Invasion": "pT2"). This shows the model that when a similar report is encountered, the model should generate a similar response. Including examples is referred to as in-context learning and it can enhance responses by guiding the model to adhere to the specifications of the expected response [18].

### 2.7.4 Uncertainty Handling

In the last component of the prompt, we have specified for the model how to handle ambiguous reports, confusing cases and reports with insufficient information. We have mentioned in the prompt that such cases exist and the model can respond with "Not Available" if it encountered one. It is crucial to provide the model the option to "not provide an answer" or it might forcibly generate an answer, potentially leading to misinformation.

## 2.8 Confidence Estimation and Rescaling

Estimation of confidence is crucial in reflecting how reliable a response is. The main goal of this work is not to make a model that is optimal and makes no mistakes, but to make a model that is able to reflect its reliability as a measurement of confidence. The confidence of the final response made by the Validator for each query field $q$ is estimated upon multiple factors as shown in Equation 1.

$$C_q = \frac{E_{Confidence} + V_{Correct} + V_{Confidence} + V_{Correction} + V_{\%Correct}}{5} \qquad (1)$$

Here, $E_{Confidence}$ corresponds to the confidence reported by the Extractor, $V_{Correct}$, $V_{Confidence}$ and $V_{Correction}$ are the confidences reported by each field of the Validator's response. In the equation, the term $V_{\%Correct}$ represents the frequency with which the Validator's response aligns with the Extractor's output, reflecting a greater confidence in the response when there is agreement between both agents. Relying on one factor only may not provide an accurate representation of the model's reliability as models often exhibit overconfidence [19]. Furthermore, consistency in responses from a single prompt does not necessarily guarantee reliability, as these models can be consistent while being incorrect. We have averaged all factors as we expected the model to reflect its incorrectness on one or more of the factors due to the distinct ways of prompting (extraction and validation).

Estimated confidence is expected to be high as LMMs tend to be overconfident and consistent. We have applied Platt's probability calibration to rescale confidence of each query field $C_q$ and make it more representative of the actual performance of the model [20]. Platt's scaling works by fitting two coefficients $(A_q, B_q)$ to the sigmoid function and use it to generate scaled confidence values to be reflective of the errors made by the model.

## 2.9 Measurement of Extraction Performance

It is expected for LMMs to make mistakes considering the complexity and the extensive context needed to perform extraction tasks. We have evaluated the model's effectiveness by examining how accurately its confidence estimates reflect its correctness. For each extracted value from a given query field in the validation set, we assigned a label of +1 or -1 for correct and incorrect extractions, respectively. We then used the model's confidence scores for these extractions to compute and plot the area under the Receiver Operating Characteristic (ROC) curve for each field [21]. This measure provides an assessment of the accuracy of the model's confidence in its extractions.

## 2.10 Measurement of Effectiveness of Confidence

We hypothesised that excluding extractions with low confidence values would enhance the accuracy of the model defined as the proportion of correct extractions of a query field over the validation set. To test this hypothesis, we calculate the percentage of extractions *rejected* for falling below a specified confidence threshold and assess the accuracy of the model over *accepted* extractions that meet or exceed this threshold. The mean performance of the model is expected to increase with increasing confidence value thresholds resulting in the model effectively abstaining from producing a response when the generated response can be expected to be incorrect.

## 2.11 Analysis of Prognostic Value of Standardised Reports

The information contained in pathology reports is crucial for medical decision-making and holds prognostic value [12]. We have performed survival analysis to confirm the prognostic value of the standardised reports. The model we have used is proposed by Alzaid et. al [22] which performs transductive survival ranking. Concordance Index (c-index) [23] was calculated to evaluate the value of the survival scores given by the model. We have plotted Kaplan-Meier (KM) survival curves [24] to visualise the capability of the standardised reports to stratify patients effectively into two distinct risk categories.

## 2.12 Analysis of Prognostic Value of Report Embeddings

We conducted another experiment to demonstrate the value of reports content by transforming the text into word embeddings which are the numerical representation of the text aiming to capture its meaning and context. Transforming text to this format is one way of handling unstructured text [25]. We have transformed text reports into 1536 dimensional embeddings using OpenAI *"text-embedding-3-small"* and performed survival analysis using the same method as the analysis for standardised reports.

## 3 RESULTS

In this section, we present both qualitative and quantitative results from the proposed approach in terms of its ability to standardise reports, quality of its extractions, effectiveness of its confidence estimation for different fields as well as the prognostic significance of reports through survival analysis.

## 3.1 Standardisation of Reports

TCGA is a comprehensive and multi-centric study with pathology reports of diverse style and format collected from various institutions. The model has shown great adaptability to this variation by interpreting a wide range of reporting types. Figure 4 shows a sample report from the TCGA COAD dataset and what type of information both agents extract. Each field is extracted with its corresponding confidence score and the final output of the model is shown following RCPath reporting proforma document. More samples can be found in supplementary material which show that the model can be an effective tool in automatic information extraction as well as standardisation of pathology reports.

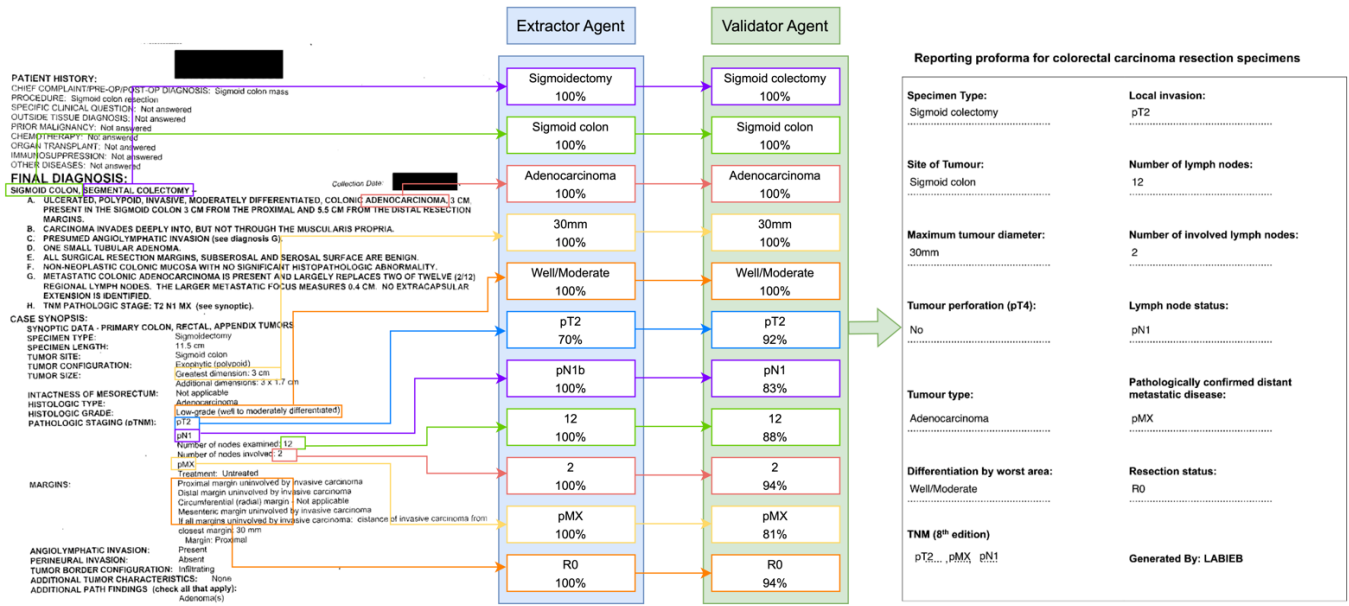

**Figure 4.** Sample of Colorectal Adenocarcinoma report and how the model produces the output. The fields are extracted from the report by the Extractor agent and passed to the Validator which returns each field with its corresponding confidence value. The final output of the model is formatted according to the standards of reporting.

## 3.2 Assessing Confidence as a Performance Indicator

In order to establish that the estimated confidence values of different fields for a given report are actually indicative of that field being extracted correctly by the model, we report the area under the ROC in Table 1 for correct extractions . As can be seen in Figure 5 (A), the confidence assignment for most fields is considerably high. Lymph node status field achieved the highest AUROC at 0.93, indicating a strong ability to reflect model performance accurately through confidence scores. For this field, the model made a larger number of extraction errors and it could be accounted to the variation in reporting subcategories. For example, a report may refer to a case with 4 metastatic nodes as the main category "pN2" instead of the subcategory "pN2a". The model was instructed to strictly follow the guidelines and report subcategories when possible. However, for some cases the model would not apply reasoning while extracting the information if it was explicitly mentioned as the main category within the text. Tumour type achieved 0.89 AUROC meaning higher confidence estimation successfully indicated correct extractions for this field. On the other hand, the model correctly extracted histologic grade in 99% of samples in the validation set but the confidence AUROC is not as good as the remaining fields. The lowest AUROC was for metastatic lymph nodes where the confidence estimation did not represent accurate extractions as effectively as in other fields which could be attributed to the field often not being explicitly mentioned in reports (e.g. 1/5 in reports means 1 metastatic node out 5 examined nodes). This high percentage of correct extractions for this field in particular is attributed to its clear definition as the response should either be 0 (low grade) or 1 (high grade) which lowers the chances of errors.

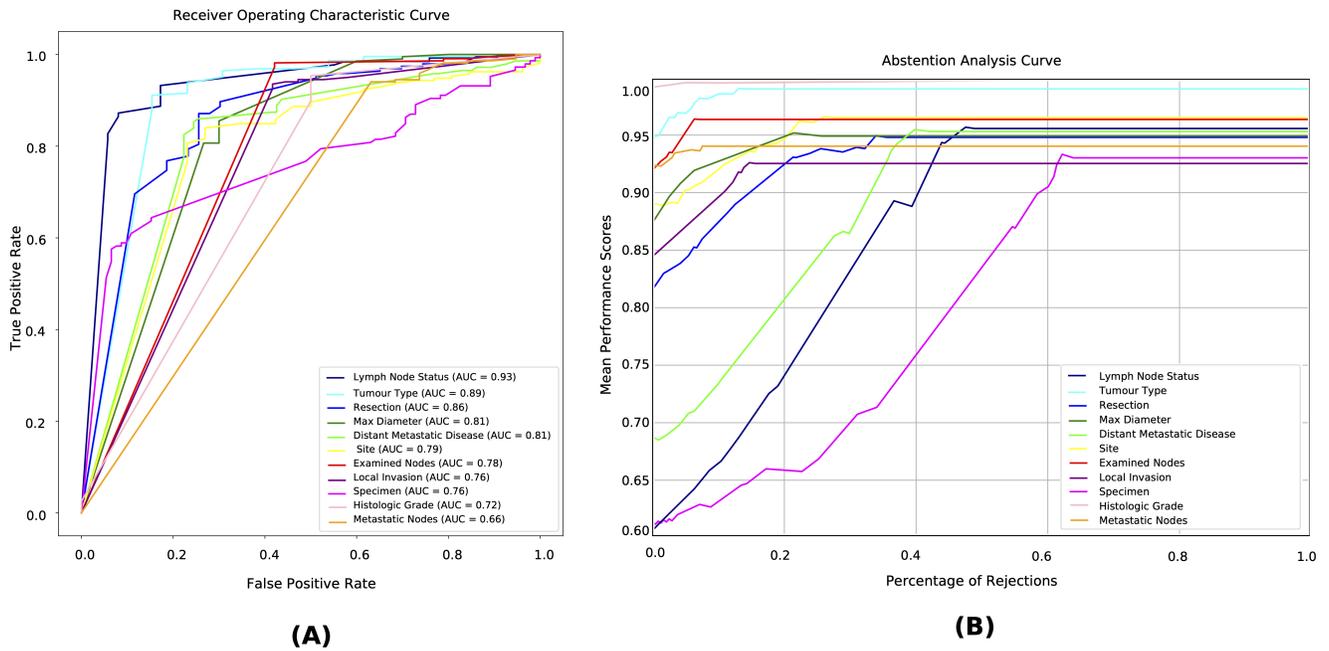

**Figure 5.** (A) Receiver Operator Curves for extracted fields plotted to measure how reflective the estimated confidence of the model's performance. AUROC for each field is shown in the legend. (B) Plot of mean performance when samples are rejected based on their confidence.

**Table 1** : AUROC scores for all extracted fields

| Field | AUROC | Field | AUROC | Field | AUROC |
| --- | --- | --- | --- | --- | --- |
| **Lymph Node Status** | 0.93 | **Distant Metastatic Disease** | 0.81 | **Specimen Type** | 0.76 |
| **Tumour Type** | 0.89 | **Site** | 0.79 | **Histologic Grade** | 0.72 |
| **Resection** | 0.86 | **Examined Nodes** | 0.78 | **Metastatic Nodes** | 0.66 |
| **Maximum Diameter** | 0.81 | **Local Invasion Status** | 0.76 | | |

## 3.3 Abstention Analysis Curves

In order to establish that the confidence values produced by the model can be used to reject those fields that are not extracted correctly, we have analysed how the accuracy of extraction changes as those fields with low confidence values are rejected. The percentage of *rejected* samples are plotted against mean performance score in Figure 5 (B). The plots clearly demonstrate a positive trend indicating that as the rejection rate increases, the model's performance improves accordingly. This suggests that selectively considering higher-confidence scores can significantly improve overall performance. Again, the trend is clearer with Lymph node status due to the increased error rate in performance scores and the variation in confidence estimates. It is less apparent in histologic grade due to the effect of the model's low extraction errors.

## 3.4 Prognostic value of pathology reports

From our survival analysis for predicting Disease Specific Survival (DSS) based on the standardised reports and the report embeddings, we have been able to stratify patients into two different risk groups. The c-index we have achieved is reasonably high for up to $0.73\pm0.04$ for embeddings and $0.74\pm0.04$ for the extracted fields of the standardised reports. This high c-index indicates that the content of the reports is a reliable predictor of survival with strong prognostic relevance in assessing patient outcomes.

We have performed risk stratification based on the predicted survival and the results are shown in the KM curves plotted in Figure 6 (A) for the report embeddings and (B) for the standardised reports. It can be noted that the survival time of the low risk group is significantly higher than the high risk group with a p-value of ($p \ll 0.005$). By analysing the reports in TCGA, we have been able to prognosticate patient survival outcomes effectively which is essential in planning proper intervention or treatment.

In addition, we have performed univariate analysis on each field in the standardised report to measure their contribution to the patient survival outcome prediction. Figure 6 (C) visualises the weight distributions for each field. Fields are ranked from top to bottom with the top having the highest weight values, thus contributing the most to the survival score generated by the survival model. The central vertical line in each field represent the median of the weight distribution. The top three fields are

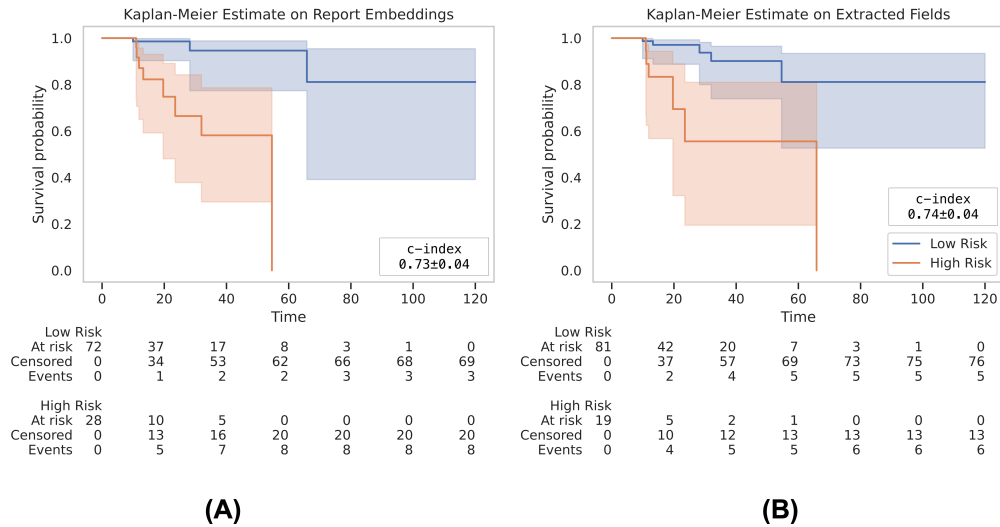

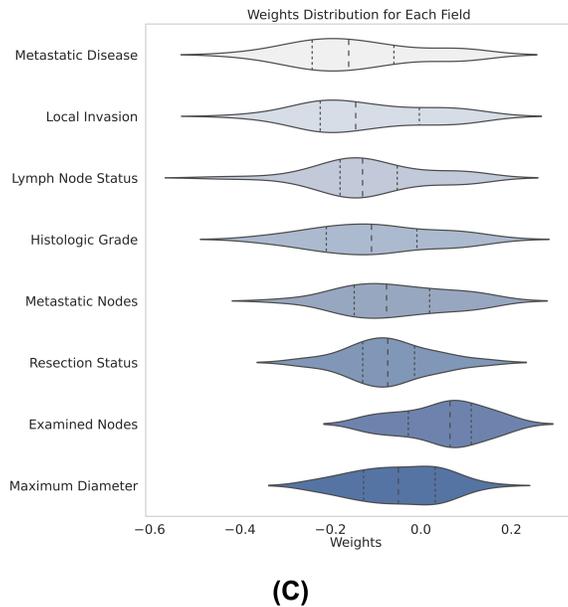

**Figure 6.** (A) Kaplan-Meier (KM) survival curves comparing high-risk and low-risk groups over time for survival probabilities derived from report embeddings, (B) displays KM curves based on extracted fields. (C) Violin plot for the distribution of weights of the standardised report fields for the survival model.

Metastatic Disease, Local Invasion, and Lymph Node Status which aligns with the TNM staging system known to describe the progression of cancer. TNM is used to determine the best treatment plan for the patient [26]. This alignment affirms the standardised reports' value and its clinical relevance in accurately predicting patient outcomes. A more detailed plot for the weights can be found in supplementary.

## 3.5 LABIEB Website

We have developed a web-page for our model to make this work accessible to everyone (https://labieb.dcs.warwick.ac.uk). The interface is intuitive and user-friendly to ensure accessibility for individuals across various levels of technical expertise. The page offers the user the option to upload the report file in either PDF or text format. The file will be processed through openAI GPT-4 Turbo API commands and the results will be displayed on the same page. The page provides multiple ways to download the results: JSON format, PDF format (follows the same template as the RCPath), and in text format.

## 4 DISCUSSION

We proposed a two-stage information extraction from pathology reports using LMM. This framework enables the estimation of a confidence score to reflect the accuracy of the extracted information to address the limitations of existing methods. Expressing uncertainty gives the user of the data the liberty to reject samples of extracted data with lower confidence. The results show that this framework is effective in estimating a confidence score that is representative of the accuracy of the extracted information. This experiment is tailored exclusively for COAD cases and might not directly apply to other cancer types. However, the underlying framework is versatile and can be adapted for any information extraction task.

Although LMMs are acknowledged for their powerful capabilities, the privacy implications of commercial models like GPT remain a concern. The handling and storage of prompts sent by users, coupled with the model's ability to remember and potentially leak this information in future responses, pose a privacy risk. Therefore, we recommend anonymising any patient-specific information -such as ID, date of birth or any other details that could lead to patient identification- before utilising any commercial LMM, including the framework integrated into our website.

This issue rises the necessity for open-source LMMs fine-tuned on pathology data that can operate locally without compromising privacy. While open-source LLMs do exist (such as Llama-2 [27] and Falcon [28]) they require further development to reach the sophistication of their commercial counterparts and are limited to text input only.

We demonstrated that GPT-4 is capable of performing information extraction in a structured format and can indicate its confidence through the proposed confidence estimation technique. Nonetheless, there remains room for improvement as the results and confidence estimations do not yet achieve the level of precision expected of a medical practitioner. Therefore, as a future direction, we propose investigating the capabilities of GPT further by fine-tuning it with pathology reports to assess its adaptability across different types of cancers.

We provided the model with sufficient context for enhanced comprehension by incorporating standard categorisation schemes and examples. Determining the right amount of context is challenging, with the primary constraint being the token window limit in API requests which dictates the maximum size of the prompt.

In conclusion, the proposed framework demonstrates promising potential and is versatile enough to be employed in numerous information extraction tasks from unstructured text. Standardising reports is expected to enhance the sharing of diagnostic and treatment information among medical professionals. From a machine learning perspective, leveraging such detailed clinical data allows for the training of more accurate and sophisticated models, enhancing healthcare delivery and patient outcomes.

## Acknowledgements

EZ is supported by the Saudi Cultural Bureau in London, UK. FM acknowledges funding from EPSRC grant EP/W02909X/1. Authors acknowledge Skiros Habib for their assistance with the deployment of the website. Authors acknowledge support from the Tissue Image Analysis (TIA) centre at the University of Warwick.

## Conflict of Interest Statement:

HE reports working ad-hoc part time sessions for Histofy Ltd (a start-up company developing artificial intelligence algorithms for digital pathology). DS is a co-founder of Histofy Ltd and reports personal fees from Royal Philips, outside the submitted work. FM reports research funding from GlaxoSmithKline outside of the scope of this work. All other authors have no conflict of interest.

## Author Contributions Statement

EZ and FM conceptualised the study. EZ conducted and evaluated the experiments with guidance of FM. GP provided feedback and advice for the experimental setup. HE and DS assisted in the validation of extracted information. All authors corrected and agreed to the final version of the manuscript.

# REFERENCES

1    Sedlakova J, Daniore P, Wintsch AH, *et al.* Challenges and best practices for digital unstructured data enrichment in health research: A systematic narrative review. *PLOS Digital Health* 2023; **2**: e0000347

2    Coden A, Savova G, Sominsky I, *et al.* Automatically extracting cancer disease characteristics from pathology reports into a Disease Knowledge Representation Model. *J Biomed Inform* 2009; **42**: 937-949

3    Yoon H-J, Ramanathan A, Tourassi G. Multi-task Deep Neural Networks for Automated Extraction of Primary Site and Laterality Information from Cancer Pathology Reports. In Advances in Big Data. Springer International Publishing, ; 195-204.

4    Yoon H-J, Roberts L, Tourassi G. Automated histologic grading from free-text pathology reports using graph-of-words features and machine learning. In 2017 IEEE EMBS International Conference on Biomedical & Health Informatics (BHI). , 2017; 369-372.

5    Wu J, Zhang R, Gong T, *et al.* BioIE: Biomedical Information Extraction with Multi-head Attention Enhanced Graph Convolutional Network. In 2021 IEEE International Conference on Bioinformatics and Biomedicine (BIBM). , 2021; 2080-2087.

6    Choi HS, Song JY, Shin KH, *et al.* Developing prompts from large language model for extracting clinical information from pathology and ultrasound reports in breast cancer. **41**: 209-216 DOI:10.3857/roj.2023.00633.

7    Stroganov O, Schedlbauer A, Lorenzen E, *et al.* Unpacking Unstructured Data: A Pilot Study on Extracting Insights from Neuropathological Reports of Parkinson's Disease Patients using Large Language Models. DOI:10.1101/2023.09.12.557252.

8    OpenAI. GPT-4 Technical Report. 2023

9    Nori H, King N, McKinney SM, *et al.* Capabilities of gpt-4 on medical challenge problems. *arXiv preprint arXiv:230313375* 2023

10   Truhn D, Loeffler CML, Müller-Franzes G, *et al.* Extracting structured information from unstructured histopathology reports using generative pre-trained transformer 4 (GPT-4). *J Pathol* 2024; **262**: 310-319

11   Mizrahi M, Kaplan G, Malkin D, *et al.* State of What Art? A Call for Multi-Prompt LLM Evaluation. 2024

12   The Royal College of Pathologists. Dataset for Histopathological Reporting of Colorectal Cancer. April 2023

13  Tesseract OCR contributors. Tesseract User Manual. 2021

14  Shi Y, Peng D, Liao W, *et al.* Exploring OCR Capabilities of GPT-4V(ision) : A Quantitative and In-depth Evaluation. 2023

15  Sclar M, Choi Y, Tsvetkov Y, *et al.* Quantifying Language Models' Sensitivity to Spurious Features in Prompt Design or: How I learned to start worrying about prompt formatting. 2023

16  Tang X, Zong Y, Phang J, *et al.* Struc-Bench: Are Large Language Models Really Good at Generating Complex Structured Data? September 2023

17  Li C, Wang J, Zhang Y, *et al.* Large Language Models Understand and Can be Enhanced by Emotional Stimuli. DOI:10.48550/arXiv.2307.11760.

18  Min S, Lyu X, Holtzman A, *et al.* Rethinking the Role of Demonstrations: What Makes In-Context Learning Work? 2022

19  Xiong M, Hu Z, Lu X, *et al.* Can LLMs Express Their Uncertainty? An Empirical Evaluation of Confidence Elicitation in LLMs. In The Twelfth International Conference on Learning Representations. , 2024.

20  Platt J. Probabilistic outputs for support vector machines and comparison to regularized likelihood methods. In Advances in Large Margin Classifiers. , 2000.

21  Bradley AP. The use of the area under the ROC curve in the evaluation of machine learning algorithms. *Pattern Recognit* 1997; **30**: 1145-1159

22  Alzaid E, Dawood M, Minhas F. A Transductive Approach to Survival Ranking for Cancer Risk Stratification. In Knowles DA, Mostafavi S, eds. Proceedings of the 18th Machine Learning in Computational Biology Meeting. Vol 240. Proceedings of Machine Learning Research. PMLR, 2024; 101-109.

23  Jr FEH, Lee KL, Califf RM, *et al.* Regression modelling strategies for improved prognostic prediction. *Stat Med* 1984; **3**: 143-152

24  Kaplan EL, Meier P. Nonparametric Estimation from Incomplete Observations. **53**: 457-481 DOI:10.1080/01621459.1958.10501452.

25  Egger R. Text Representations and Word Embeddings. In Egger R, ed. Applied Data Science in Tourism: Interdisciplinary Approaches, Methodologies, and Applications. Cham: Springer International Publishing, 2022; 335-361.

26  Cancer Staging | Has Cancer Spread | Cancer Prognosis | American Cancer Society. [Accessed April 25, 2024] Available from: https://www.cancer.org/cancer/diagnosis-staging/staging.html.

27  Touvron H, Martin L, Stone K, *et al.* Llama 2: Open Foundation and Fine-Tuned Chat Models. 2023

28  Almazrouei E, Alobeidli H, Alshamsi A, *et al.* The Falcon Series of Language Models: Towards Open Frontier Models. 2023

29  Chen J, Lin H, Han X, *et al.* Benchmarking Large Language Models in Retrieval-Augmented Generation. 2023